\documentclass[runningheads]{llncs}
\usepackage[T1]{fontenc}
\usepackage{graphicx}
\usepackage{color}
\usepackage{amsmath}
\usepackage{bbm}
\usepackage[noend,ruled,linesnumbered]{algorithm2e}
 \usepackage{multirow}

\newcommand{\cf}{\emph{cf.}~}

\makeatletter
\newcommand{\printfnsymbol}[1]{%
  \textsuperscript{\@fnsymbol{#1}}%
}
\makeatother

\begin{document}
\title{Video-based Surgical Skills Assessment using Long term Tool Tracking}
%
%
\author{Mona Fathollahi\thanks{equal contribution}\inst{1}, Mohammad Hasan Sarhan\printfnsymbol{1}\inst{2}, Ramon Pena\inst{3}, Lela DiMonte\inst{1}, Anshu Gupta\inst{3}, Aishani Ataliwala\inst{4}, Jocelyn Barker\inst{1}}
%
\authorrunning{M. Fathollahi, M. H. Sarhan, et al.}
%
\institute{Johnson \& Johnson Santa Clara, CA, USA \and
Johnson \& Johnson Medical GmbH, Hamburg, Germany \and
Johnson \& Johnson Raritan, NJ, USA \and
Johnson \& Johnson Seattle, WA, USA\\
\email{\{mfatholl,msarhan\}@its.jnj.com}}

\maketitle              
\begin{abstract}
Mastering the technical skills required to perform surgery is an extremely challenging task. Video-based assessment allows surgeons to receive feedback on their technical skills to facilitate learning and development. Currently, this feedback comes primarily from manual video review, which is time-intensive and limits the feasibility of tracking a surgeon's progress over many cases. In this work, we introduce a motion-based approach to automatically assess surgical skills from surgical case video feed. The proposed pipeline first tracks surgical tools reliably to create motion trajectories and then uses those trajectories to predict surgeon technical skill levels. The tracking algorithm employs a simple yet effective re-identification module that improves ID-switch compared to other state-of-the-art methods. This is critical for creating reliable tool trajectories when instruments regularly move on- and off-screen or are periodically obscured. The motion-based classification model employs a state-of-the-art self-attention transformer network to capture short- and long-term motion patterns that are essential for skill evaluation. The proposed method is evaluated on an in-vivo (Cholec80) dataset where an expert-rated GOALS skill assessment of the Calot Triangle Dissection is used as a quantitative skill measure. We compare transformer-based skill assessment with traditional machine learning approaches using the proposed and state-of-the-art tracking. Our result suggests that using motion trajectories from reliable tracking methods is beneficial for assessing surgeon skills based solely on video streams.

\end{abstract}
\section{Background and Introduction}


Tracking a surgeon's instrument movements throughout a minimally invasive procedure (MIP) is a critical step towards automating the measurement of a surgeon's technical skills. Tool trajectory-based metrics have been shown to correlate with surgeon experience~\cite{Fard2017}, learning curve progression~\cite{Shafiei2015}, and patient outcome measures~\cite{Hung2018.1}. Typically, it is only feasible to calculate these metrics for training simulators or robot-assisted MIP procedures directly from the robot kinematic output. Since most MIP cases worldwide are performed laparoscopically rather than robotically, a broadly applicable solution is needed for instrument tracking. Using computer vision techniques to generate tool positional data would enable calculation of the motion metrics based on surgical case video alone and not rely on data outputs of a specific surgical device.

A Surgeon's technical skills are typically evaluated during video review by experts, using rating scales that assign numerical values to specific characteristics exhibited by the surgeon's movements. The Global Operative Assessment of Laparoscopic Skills (GOALS) scale uses a five-point Likert scale to evaluate a surgeon's depth perception, bimanual dexterity, efficiency, and tissue handling~\cite{Vassiliou2005}. GOALS scores are sometimes used as a measure for which to compare other methods of surgeon technical skill classification~\cite{Dubin2018}.
Instead of relying on manual review, many studies have looked at methods to automate the evaluation of technical skills based on a surgeon's movement patterns~\cite{Levin2019}. Metrics of interest such as path length, velocity, acceleration, turning angle, curvature, and tortuosity can be calculated and used to distinguish surgeon skill level~\cite{Fard2018}. Motion smoothness metrics such as jerk have been shown to indicate a lack of mastery of a task, indicating a surgeon's progression along the learning curve~\cite{Azari2019}.

Tool trajectory-based metrics have been calculated using robotic system kinematic output~\cite{DeWitte2021,Zia2018,Rivas-Blanco2021}. Estimating instrument trajectories from the laparoscope video feed using image-based tracking can unlock the ability to further explore automated assessment of technical skills, as no extra sensors or robotic system access are needed. Metrics are then calculated directly from the image-based tracking results~\cite{Law2020,Lee2020,Jin_2018_8354185,ganni2020validation,perez2016construct}. However, there are limitations to these approaches that we try to address in this work. For example, a surgical tool detection model is trained in~\cite{Jin_2018_8354185} to estimate instrument usage timelines and tool trajectory maps per video. But skill assessment step is not automatic, and each visualization is qualitatively evaluated by a human to estimate surgeon skill. Third-party software in semi-automatic mode is used in~\cite{ganni2020validation} to track tools. This is not a feasible solution for processing thousands of videos. The duration of each video in their dataset is 15 seconds and the dataset is not publicly released. Some approaches register 3D tool motion by employing extra sensors or a special lab setting. For example,~\cite{perez2016construct} use two cameras in an orthogonal configuration in addition to marking the distal end of surgical tools with colored tapes to make them recognizable in video feeds.

\par Image-based surgical instrument tracking is an attractive solution to the temporal localization problem but requires robustness towards variations in the surgical scene which may lead to missed detections or track identity switches. Approaches achieve state-of-the-art performance on public tracking benchmarks such as MOTChallenges~\cite{MOT16} and OTB datasets~\cite{wu2013online}. However, when it comes to real-world or long videos, most of them have several shortcomings because they are designed for evaluating short-term tracking methods. For example, the longest sequence in the MOT17 and OTB datasets is only 30 and 129 seconds respectively. To alleviate these problems and obtain suitable motion features for skill assessment, we propose a video-based framework for extracting motion tracks from surgical videos reliably and using them in surgeon skill assessment. The framework utilizes a novel tracking algorithm that is suitable for long sequence tracking with a minimal amount of identity switches. This is achieved with a tailored cost criterion that takes semantic, spatial, and class similarity into consideration when assigning detections to tracklets. The reliable motion features are utilized in assessing surgeon efficiency in laparoscopic cholecystectomy segment of calot triangle dissection. We compare the tracking method with a state-of-the-art tracking method and validate the motion-based skill assessment using feature extraction and learning-based transformer approaches.

\section{Materials and Methods}
\subsection{Dataset Description}
An open-source dataset, Cholec80~\cite{Twindada2017}, was used to perform this analysis. The Cholec80 dataset consists of laparoscopic cholecystectomy surgical case videos performed by 13 surgeons, and are labeled with phase annotations. The Calot Triangle Dissection phase was chosen to be evaluated as it showcases a surgeon's fine dissection skills. To account for the variance of the duration of this phase in different videos, we only consider the last three minutes of Calot Triangle Dissection for all our experiments.
To evaluate our tracking algorithm, 15 videos of cholec80 dataset have been annotated at half the original temporal resolution. The rest of the videos were used for training the detection and reIdentification models. In addition, to evaluate our Skill assessment model, two expert surgeons annotate the Calot Triangle Dissection segment of all 80 videos of Cholec80 dataset using the GOALS assessment tool from 1 to 5. The GOALS efficiency category in particular was chosen to be representative of surgeon movement patterns that relate to technical skill progression. Therefore, efficiency scores from the two experts were averaged for each case. We decided to binarize the averaged score due to the limited size of the dataset. With 5 classes, the individual class representation was too small for learning. The cutoff point (3.5) is selected as it shows the best agreement between the annotators. Using this cutoff threshold, 29 cases belong to the low-performing group and the other 51 in the high-performing group. The performance may degrade on different thresholds as the agreement between annotators is lower.

\subsection{Tracking Algorithm}
\label{sec:tracking}
Our proposed tracking algorithm belongs to the Tracking-by-detection category to solve the problem of tracking multiple objects. Tracking-by-detection algorithms consist of two steps: (i) objects in the frame are detected independent of other frames. (ii) Tracks are created by linking the corresponding detections across time. A corst or similarity score is calculated for every pair of new detections and active tracks to match new data to tracks through data association. When a detection can not be matched with active tracks, either due to the emergence of a new object or a low similarity score, a new track is initialized. Regarding the tracking, we have two main contributions: 1) We propose a new cost function and 2) a new policy in track recovery.

The proposed tracking algorithm is summarized in Algorithm~\ref{alg:cap}. At each frame, the detection algorithm detects all tools in the frame.  We utilize yolov5~\cite{glenn_jocher_yolov5} that is trained on a subset of CholecT50~\cite{innocent2021rendezvous}. We have annotated the dataset in-house with the bounding box location and class of each surgical instrument present in the scene. This resulted in more than 87k annotated frames and 133k annotated bounding boxes. We use a pre-trained model on the COCO dataset and freeze the model backbone (first 10 layers) while training. Each detection is passed through a re-Identification network to extract appearance features. Furthermore, for each active track, the locations of the corresponding tool are predicted using a Kalman filter. This information is used to construct a cost matrix that is fed to the Hungarian algorithm to assign detections to tracks. The next step is "track recovery" which is mainly matching unassigned detections in the previous step to some inactive tracks if it is possible to do so.

\begin{algorithm}[!ht]
 \caption{Tracking Aglorithm}\label{alg:cap}
\ForEach{$\text{frame}\, (i)$}
{
  $detected\_tools \gets \text{detect tools in frame}(i)$\;
  \ForEach {$d \in detected\_tools$}
  {
    $appearance\_features[d] \gets \text{extract appearance features}(d)$\;
  }
  \ForEach {$t \in active\_tracks$}
  {
    \ForEach {$d \in detected\_tools$}
    {
      $cost\_matrix[t, d] \gets cost ((t, d))$\;
    }
  }
  $assignments \gets \text{run hungrian assignment}(cost\_matrix)$\;

  \ForEach {$d \in unassigned\_detections$}
  {
    \eIf{$d\, \text{is matchable to an inactive track}$}
    {
     recover inacive track\;
    }{
     initialize a new track using $d$\;
    }
  }
  $\text{update/inactive tracks}(assignments)$\;
}
\end{algorithm}

\subsubsection{Re-Identification Network}
The goal of the reidentification network is to describe each image from a tool track with a feature embedding such that the feature distance between any two crops in a single track (a) or (b) is small, while the distance of a pair that is made out of track (a) and (b) is large comparatively. To accomplish this goal, we train a TriNet~\cite{hermans2017} architecture to represent each tool bounding box with a 128-dimensional vector.

\subsubsection{Cost Function Definition}

To be able to associate detected objects with the tracks, first a cost matrix between each active track, $t$, and new detection $d$ is constructed. Next, the cost matrix is minimized using the Hungarian algorithm~\cite{kuhn1955}. The cost function is defined as equation~\ref{cost} which is a combination of appearance feature, detection, track ID matching, and spatial distance, where $t$ is a track, $d$ is a detection and $D(.,.)$ is the distance function.
\begin{align}
\label{cost}
cost(t,d) = D_{feat}(t,d)+ M.\mathbbm{1}_{D_{spatial}(t,d)>\lambda_{sp}} + M.\mathbbm{1}_{d.DetClass\neq t.classID}
\end{align}

In this equation, the first term represents the dissimilarity between the appearance re-ID feature of track $t$ and detection $d$. To achieve faster tracking and adapt to the appearance changes of a track across frames, we employed short-term re-identification~\cite{bergmann2019} where only the  N recent frames of a track are used to calculate the distance to a new detection in the learned embedding space. The second term is the spatial distance between the center of detection, $d$, and the predicted Kalman position of the track, $t$. We assume that surgical tools have moved only slightly between frames, which is usually ensured by high frame rates, 25 fps in our case. If this distance is greater than the $\lambda_{sp}$ threshold, then the corresponding element in the cost matrix is given a very high number to prohibit this assignment. The last term is referred to 'classID' match. This term adds a high bias, $M=1000$, to the cost if the predicted class label of detection does not match the class label of the object presented in the track.
\subsubsection{Track Recovery}
In the data association step, any detection that can not be assigned to an active track is fed into the Track Recovery submodule. In this submodule, the cost of assigning an unassigned detection $d$ to any inactive track $t$ is calculated as follows:
\begin{align}
cost(t, d) = D_{feat}(t,d) . M.\mathbbm{1}_{D_{spatial}(t,d)>\lambda_{sp} \land d.DetClass\neq t.classID}
\end{align}

This step helps to re-identify the tool that becomes occluded or moves out of the screen. The intuition behind this cost function is that, if the tool emerges from an occluded organ, the detection algorithm usually fails to correctly identify its classID, while their spatial distance can be reliably used to match them.

\subsection{Feature Based Skill Assessment}
\label{sec: skill}
Trajectories from the tracking algorithm can be decomposed into coordinates in the 2D pixel space of the video feed. These trajectories and their variations over time can then be used in the calculation of motion metrics that describe the path of the tool. In our feature-based skill assessment, metrics from the literature~\cite{oropesa2013relevance,estrada2014development} including distance (path length), velocity, acceleration, jerk, curvature, tortuosity, turning angle, and motion ratio are calculated between each time step. Motion ratio was simply defined as the ratio of time the tool is in motion to the total time of the task.  These features were used in a random forest model to classify surgeons into the high and low efficiency classes.

\subsection{Learning Based Skill Assessment}
\begin{figure}[t]
\centering
\includegraphics[width=1.0\textwidth]{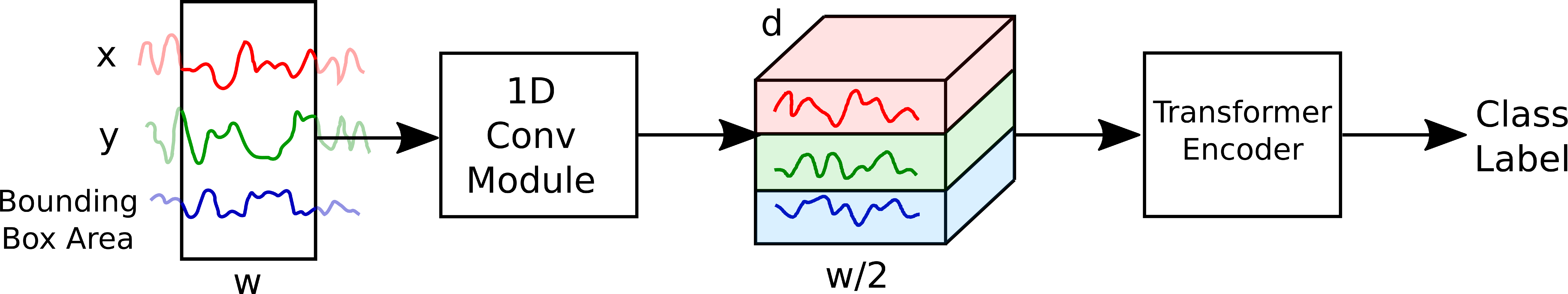}
\caption{Transformer-based model to learn surgical skill.}
\label{fig:trans-model}
\end{figure}

We compare feature-based skill assessment classification with a learning-based approach. The motion features (i.e location and bounding box area information) from the tracking method are used as inputs to the learning model and the model weights are learned by optimizing a cross-entropy loss function to classify surgeon skills using a predefined skill criterion. A transformer model~\cite{vaswani2017attention} was used due to prior excellent performance across a wide range of time-series applications~\cite{wen2022transformers}. Both short and long-range relationships of tool location are needed to get an insight quality of the surgeon's motion which is enabled by the self-attention mechanism. Fig.~\ref{fig:trans-model} shows our network. 3D tracking time series, x, y and bounding box area, is first fed through a convolutional module with BatchNormalization and Relu after each convolution: $Conv1D(d=128,k=11,s=1) + Conv1D(d=128,k=3,s=2)$ where kernel size, $k$, is the width in number of time-steps, $d$ is num of output channel and $s$ is stride. The rationale behind this module is to learn some atomic motion from tool trajectory in addition to reducing the temporal resolution of the input. The the second module is a transformer encoder with $num\_heads=7, num\_layers=2, dim\_feedforward=56$. The output of this model is the binary prediction of the efficiency of the surgeon which serves as a proxy for the surgeon's skill. To our knowledge, this is the first work that uses a transformer-based framework to classify surgeon skills from dominant surgical tools in a video sequence. Moreover, we build a multi-channel 1-dimensional convolution model for skill assessment based on inception-v3~\cite{szegedy2016rethinking}. The model shares the same pre-inception convolution blocks and the first inception block of~\cite{szegedy2016rethinking} all changed to 1D operations rather than 2D.

\section{Results}
\subsection{Tracking Model Performance}
This section compares a state-of-the-art tracking method (ByteTrack~\cite{zhang2021bytetrack}) with the proposed tracking (\cf~Sec.~\ref{sec:tracking}). ByteTrack is a simple yet effective tracking method that utilizes high confidence and low confidence detection to create and update the tracks.
Both trackers use the same detector for proposing bounding boxes and are evaluated on 15 video segments of Cholec80 dataset that were annotated at half the original temporal resolution. The results of the trackers are shown in Table~\ref{tbl:tracking}. While the proposed re-identification module has less multi-object tracking accuracy ($3\%$ relative decline), it greatly reduces the identity switches ($141\%$ relative improvement) which is an important factor in tracking surgical instruments. It is worth noting the higher number of false negatives in the re-identification module is due to not using the lower confidence bounding boxes in the tracking which mainly come from the assistant tools as they appear partially in the scene. This should not hinder the skill assessment performance as we use the longest paths motion data for assessing surgeon skills.
\begin{table}
\centering
\caption{Tracking results of ByteTrack and our proposed tracker.}
\label{tbl:tracking}
\begin{tabular}{l|ccccccc}
\hline
Method            & IDs $\downarrow$ & MOTA $\uparrow$  & FP $\downarrow$  & FN $\downarrow$  & Precision $\uparrow$ & Recall $\uparrow$ & \#Objects \\ \hline
ByteTrack         & 210 & 89.2\% & 2214 & 2615 & 95.2\%    & 94.4\% & 46479     \\
Proposed tracking & 87  & 86.6\% & 783  & 5367 & 98.1\%      & 88.5\%   & 46479     \\ \hline
\end{tabular}
\end{table}

\subsection{Skill Assessment on Cholec80}

\paragraph{Data preprocessing and augmentation:}
The Cholec80 data contains sequences of real in-vivo surgeries meaning. The number of present tools is not guaranteed as tools may leave and come back to the scene, or the camera might be out of the body for a few frames creating missing data points in the motion sequences. We focus on the tool that is present the most in the frames of the last 3 minutes of the Calot's triangle dissection step assuming this is the main tool in the scene that is performing the dissection. We create a temporal mask for motion features of the main tool that is true in the time stamps where the tool is present in the scene and false otherwise to utilize the presence information of the main tool. Lastly, we use a fixed window of motion features for each experiment and randomly change this window in different training iterations, enabling models that are invariant to the length of the task and more robust towards variations of unseen data. In all experiments, we use the same dataset with 5-fold cross-validation. To augment the training dataset, we randomly choose a different temporal window of the length of 700 (28s) out of 3 minutes duration of the target surgical phase.

\paragraph{Feature-based skill assessment:}
Results from the feature-based skill assessment classification models indicate very low reliability between motion metrics and performance scores. As indicated in Table~\ref{tbl:cholec80}, all the model assessment metrics, i.e. precision, recall, accuracy, and kappa were generally below acceptable standards. This difference from prior literature is likely due to the challenges of working with 2D computer-vision-based tracks compared to 3D robot kinematic sensor data.

\paragraph{Learning-based skill assessment:}
Learning skill directly from motion allows automatic extraction of features that are more suitable for the task. We run a Bayesian hyper-parameter search of learning-based models and show the results of the best-performing models. The data is normalized before it is fed to the model and random oversampling is employed to mitigate class imbalance problems. The learning-based methods outperformed manual feature extraction methods which suggests that the learning process is able to extract features that are more meaningful for the efficiency calculation.

\begin{table}
    \centering
    \caption{Skill assessment results on Cholec80 dataset.}
    \label{tbl:cholec80}

    \begin{tabular}{l|l|ccccc}
        \hline
        \multirow{2}{*}{Method} & \multirow{2}{*}{Tracking} & \multicolumn{5}{c}{Efficiency}        \\ \cline{3-7}
                                           &                           & Precision & Recall & Acuuracy & Kappa & p-value \\ \hline
        Feature-based                      & Proposed                  & 0.68      & 0.52   & 0.65     & 0.30     & 0.0090320  \\
        1D Convolution                     & Proposed                  & 0.83      & 0.75   & 0.74     & 0.45     & 0.0382200  \\
        Transformer                        & ByteTrack                 & 0.73      & 0.73   & 0.69     & 0.36     & 0.0483700   \\
        Transformer                        & Proposed                  & \textbf{0.88}      & \textbf{0.84}   & \textbf{0.83}     & \textbf{0.63}         & \textbf{0.0001962}      \\ \hline
    \end{tabular}

\end{table}

\section{Discussion}
Automated surgical skill assessment has the potential to improve surgical
practice by providing scalable feedback to surgeons as they develop their
practice.  Current manual methods are limited by the subjectivity of the
metrics resulting in the need for multiple evaluators to obtain reproducible
results. Here our method demonstrated the ability to agree with the
consensus of our two annotators to a greater degree than the individual
annotators agreed with each other (Kappa 0.63 model vs 0.41
annotators). This indicates the classification created by the model is
comparable to human-level performance.

A key contributor is the ability of our model to track objects over long temporal periods.  Our model showed a large reduction in
the number of ID switches compared to the state of the art, meaning objects
could be continuously tracked for much longer periods than with prior
methods.  This ability to consistently track the tools resulted in an
improved ability to classify surgeon performance.

Another novel contribution of this work is the ability to classify skills
directly from the tool tracking information.  Prior work has shown the
ability to use feature-based approaches to distinguish surgeons of differing skill levels~\cite{Zia2018}.  However, these approaches have generally been performed in lab tasks or using kinematic data where tools cannot be obfuscated and the scale of the motion is standardized.  In procedure video data, tools may frequently be not visible to the camera and the scale of the motion is highly dependent on where and how the camera is moved and placed during the operation.  In these more realistic scenarios, we found a feature-based approach to be much less successful at classifying surgeon skills than directly classifying based on the tracked motion.

One limitation of this work is that to standardize our inputs we
only assessed the end of the Dissection of Triangle of Calot surgical step.
Future studies can determine how well this method will generalize to other dissection steps or even other types of surgical actions such as suturing or stapling.  Additionally, we would like to explore this method's ability to assess other qualities of surgeon action such as bimanual dexterity and depth perception.

\subsubsection{Acknowledgements} We would like to thank Derek Peyton for orchestrating annotation collection for this project.

%
%
%
\bibliographystyle{splncs04}
\bibliography{bibliography}






\end{document}